\documentclass{article} 
\usepackage{iclr2016_conference,times}
\usepackage{hyperref}
\usepackage{url}
\usepackage{amsfonts}
\usepackage{amsmath}
\usepackage{amssymb}
\usepackage{bm}
\usepackage{graphicx}
\usepackage{subcaption}
\usepackage{comment}
\usepackage{color, soul}

\title{Learning Representations Using \\ Complex-Valued Nets}

\author{Andy M. Sarroff\textsuperscript{1}, Victor Shepardson\textsuperscript{2} \& Michael A. Casey\textsuperscript{1,2}\\
Departments of Computer Science\textsuperscript{1} and Music\textsuperscript{2}\\
Dartmouth College\\
Hanover, NH 03755, USA \\
\texttt{sarroff@cs.dartmouth.edu} \\
}

\begin{document}

\maketitle

\begin{abstract}
Complex-valued neural networks (CVNNs) are an emerging field of research in neural networks due to their potential representational properties for audio, image, and physiological signals. It is common in signal processing to transform sequences of real values to the complex domain via a set of complex basis functions, such as the Fourier transform. We show how CVNNs can be used to learn complex representations of real valued time-series data. We present methods and results using a framework that can compose holomorphic and non-holomorphic functions in a multi-layer network using a theoretical result called the Wirtinger derivative. We test our methods on a representation learning task for real-valued signals, recurrent complex-valued networks and their real-valued counterparts. Our results show that recurrent complex-valued networks can perform as well as their real-valued counterparts while learning filters that are representative of the domain of the data.
\end{abstract}


\section{Introduction}
There are many types of data for which complex-valued representations are natural and appropriate. For example wind measurements may use complex-valued data to represent joint measurements of magnitude and direction~\citep{goh2006a}. Direction of arrival is naturally modeled in ultrawideband communications using complex values~\citep{terabayashi2014a}. It is also common to work with certain real-valued data, such as audio and EEG recordings, by first transforming them to complex-valued data in the frequency domain via a complex basis, as with the Fourier transform. Motivations behind complex-valued nets (CVNNs) are that they could be used with such real-to-complex transformed data, or that they may be used for learning complex-valued representations as alternatives to Fourier and related transforms.

Whilst research into CVNNs has developed in parallel with real-valued networks, there has been relatively little focus on CVNNs in deep learning and complex-valued representation learning. Most research targets highly-specific signal-processing domains such as communications and adaptive array processing. Several factors contribute to the slow adoption of CVNNs in applications outside of these domains: first, they are difficult to train because complex-valued activation functions cannot be simultaneously bounded and complex-differentiable; second, there are few if any known methods for regularization and hyper-parameter optimization specifically developed for CVNNs. Despite such obstacles, research on CVNNs is growing steadily, with new theoretical results~\citep{zimmermann2011a,sorber2012a,hirose2012a} appearing on the heels of comprehensive treatments in recent texts~\citep{hirose2006a,mandic2009a,hirose2013a}.

Research on complex-valued activation functions and calculation of their derivatives for application to CVNNs is generally split between those composed exclusively from holomorphic activation functions and those composed exclusively from non-holomorphic activation functions. Holomorphic functions are complex differentiable at every point in a neighborhood of their domain. Non-holomorphic functions are not complex differentiable, but may be differentiable with respect to their real and imaginary parts. Each has advantages: Holomorphic activation functions may be more successful at jointly modeling phase and amplitude, however they are unbounded and therefore non-holomorphic activation functions may be preferred at times. For example if one were to build a complex-valued Long-Short Term Memory network, the only suitable gating function in the complex domain would necessarily be non-holomorphic.

In this paper, we follow a more general framework~\citep{amin2011a,amin2013a} for building CVNNs, both deep and temporal, that allows for activation functions that are composed from combinations of both holomorphic and non-holomorphic functions. We do this by utilizing the mathematical conveniences of the Wirtinger derivative, which simplifies many of the computations that are required for gradient descent for complex-valued functions.

The remainder of this paper is organized as follows. In Section~\ref{sec:back} we cover the background for CVNNs. Section~\ref{sec:wirt} describes the Wirtinger derivative and how it is applied to back-propagation for gradient descent with complex-valued activation functions. We present experiments that illustrate the utility of the methods in Section~\ref{sec:expe} and we provide concluding remarks in Section~\ref{sec:concl}.

\section{Background}\label{sec:back}
Complex numbers extend the concept of one-dimensional real numbers to two dimensions by expressing an ordered pair $(x,y)\in\mathbb{R}$ as a point $z\in\mathbb{C}$ in the complex plane, where $z = x+iy$ and $i=\sqrt{-1}$. Numbers in the complex domain provide a natural means for jointly expressing magnitude, $|z|$, and phase or direction, $\arg(z)$.

Suppose we wish to learn a function $f:\mathbb{C}^m\to\mathbb{C}^n$ by optimizing the squared error
\begin{equation}\label{eqn:obje}
\mathcal{L}(z) = |z|^2 = z\overline{z}\quad,
\end{equation}
where $(\overline{\cdot})$ denotes the complex conjugate operator. Note that $z\overline{z}=(x+iy)(x-iy)=x^2+y^2\in\mathbb{R}$ and therefore the objective function is real-valued even though $z$ is complex-valued.

Real-valued functions of complex variables are non-holomorphic and therefore their complex derivative is undefined. However if we denote $\mathcal{L}(z)=u(x,y) + iv(x,y)$ with $u:\mathbb{R}\to\mathbb{R}$ and $v:\mathbb{R}\to\mathbb{R}$ and $u$ and $v$ are real-analytic ($u$ and $v$ are differentiable) functions then it is possible to find a stationary point in the objective function. Stated more simply, we may perform gradient descent with a real-valued cost function of complex variables even though the function does not have a complex derivative.

In this paper we apply the Wirtinger derivative~\citep{wirtinger1927a} to compute the gradient~\citep{brandwood1983a}. Doing so allows us to perform differentiation on functions that are not complex-analytic but are real-analytic. It also provides a means for easily composing a combination of holomorphic and non-holomorphic functions within the computational graph of a neural network. Finally, by taking advantage of basic properties of the Wirtinger derivative, we perform gradient descent using two Jacobians per computational node.

Due to space limitations the following summary is necessarily brief. A great overview of the core mechanics of complex-valued nets and the Wirtinger derivative is found in~\citet{mandic2009a}. This and other literature are built on the theory developed in~\citet{brandwood1983a} and \citet{bos1994a} for optimization of complex-valued nets using respectively first- and second-order derivatives with Wirtinger calculus. For a deeper discussion of Wirtinger calculus and optimization techniques we refer the reader to \citet{kreutz-delgado2009a,li2008a}. Finally~\citet{amin2011a,amin2013a} advocate a framework for composing holomorphic and non-holomorphic functions in complex-valued nets.

\subsection{Complex-Valued and Real-Valued Nets}
The components of a complex-valued number can be represented as a bivariate real number, so it is natural to ask why a complex-valued representation may be preferred. A multiplication of values in the real domain yields scaling. A multiplication of complex values yields scaling and rotation. Hence if we wish to model magnitude and phase jointly, it may be more natural to do so by using a complex representation.

There are cases when we may wish to model real-valued processes in the complex domain. For instance one cannot determine the instantaneous frequency or amplitude of a real-valued periodic waveform from a single sample. Applying the Hilbert transform yields a complex-valued waveform with the same positive frequency components. However we suggest that complex-valued networks may also learn important relationships on instantaneous frequency and amplitude.

\subsection{Activation Functions}
Activation functions that are bounded and differentiable are generally desirable for training neural networks. (The rectified linear unit is a notable exception for boundedness.) Due to Liouville's theorem, the only entire (holomorphic over the entire complex domain) function that is bounded is a constant. Thus we must choose between boundedness and differentiability for complex nets.

Split-complex activation functions operate on the real and imaginary or phase and magnitude components independently and merge the outputs together. Such functions are not holomorphic. However it is easy to define a bounded split-complex activation function, for example Geourgiou and Koutsougeras' magnitude squashing activation function~\citep{georgiou1992a}. It is suggested by~\citet{mandic2009a} that the split phase-magnitude and real-imaginary approaches are appropriate when we can assume rotational or cartesian symmetry of the data, respectively.

\begin{figure}
\begin{center}
\begin{subfigure}{0.22\textwidth}
    \includegraphics[width=\textwidth]{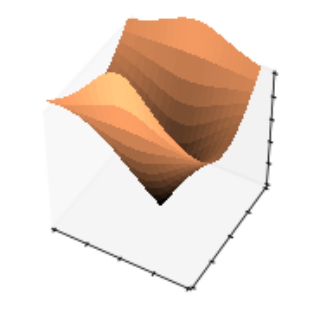}
    \caption{$|\tanh(z)|$}
\end{subfigure}
\begin{subfigure}{0.22\textwidth}
    \includegraphics[width=\textwidth]{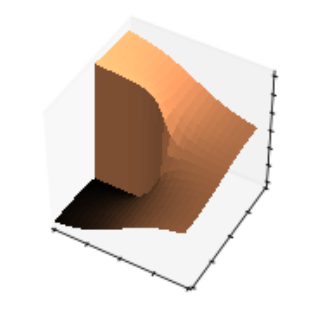}
    \caption{$\arg(\tanh(z))$}
\end{subfigure}
\begin{subfigure}{0.22\textwidth}
    \includegraphics[width=\textwidth]{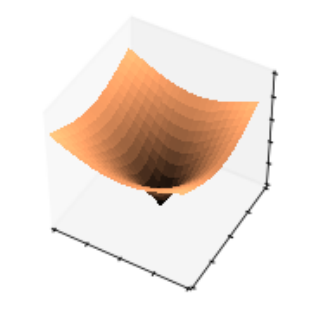}
    \caption{$\left|\frac{z}{1+|z|}\right|$}
\end{subfigure}
\begin{subfigure}{0.22\textwidth}
    \includegraphics[width=\textwidth]{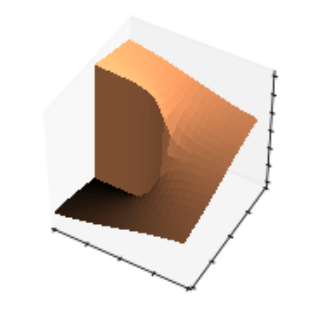}
    \caption{$\arg\left(\frac{z}{1+|z|}\right)$}
\end{subfigure}
\caption{Fully complex elementary transcendental function (a,b)~\citep{kim2003a} and split phase-magnitude (c,d)~\citep{georgiou1992a} activation functions. Axes: $\Re(z), \Im(z), f(z)$.}\label{fig:act}
\end{center}
\end{figure}

Alternatively we may choose to use fully complex activation functions that are bounded almost everywhere. Certain elementary transcendental functions have been identified which provide squashing-type nonlinear discrimination with well defined first-order derivatives~\citep{kim2003a}. These functions have singularities, but with proper treatment of weights or using other regularization mechanisms singularities may be avoided. Figure~\ref{fig:act} shows magnitude and phase surface plots for a complex $\tanh$ activation function and Geourgiou and Koutsougeras' activation function. The $\tanh$ activation has regularly spaced singularities along the imaginary axis beyond the limits of the plots.

\section{Wirtinger Framework for Gradient Descent}\label{sec:wirt}
This section outlines the routine for optimizing an arbitrary complex-valued neural network using the Wirtinger derivative and gradient descent. The network has a real-valued objective function of complex variables. It may have any combination of holomorphic and non-holomorphic activation functions. Wirtinger calculus (also known as $\mathbb{CR}$ Calculus in some texts) facilitates defining a computational graph that can be modularized as in many popular deep learning libraries, thereby allowing the construction of deep or temporal networks having many layers.

In the following subsection, the core concepts of Wirtinger calculus are reviewed. The following subsection describes the framework for building a computational graph and performing gradient descent.

\subsection{Wirtinger Derivatives}
Define $z\in\mathbb{C}$ and $x,y\in\mathbb{R}$ with $f(z)=g(x,y)=u(x,y) + iv(x,y)$. We extend the definition of $f$ to include the complex conjugate of its input variable so that
\begin{align}
f(z)&=f(z,\overline{z})=g(x,y)=u(x,y) + iv(x,y)\notag\\
z&=x+iy\notag\\
\overline{z}&=x-iy
\end{align}
Using this definition, the $\mathbb{R}$-derivative and $\overline{\mathbb{R}}$-derivative of $f$ are defined as:
\begin{align}
\left.\frac{\partial f}{\partial z}\right|_{\overline{z}\text{ is constant}}\quad\text{and}\quad\left.\frac{\partial f}{\partial\overline{z}}\right|_{z\text{ is constant}}
\end{align}
We note that the $\mathbb{R}$-derivative and $\overline{\mathbb{R}}$-derivative are formalisms, as $z$ cannot be independent of $\overline{z}$. However we treat one as constant when computing the derivative of other, applying the normal rules of calculus. Using these definitions, \citet{brandwood1983a} shows that 
\begin{align}
\frac{\partial f}{\partial z}=\frac{1}{2}\left(\frac{\partial f}{\partial x}-i\frac{\partial f}{\partial y}\right)\quad\text{and}\quad\frac{\partial f}{\partial\overline{z}}=\frac{1}{2}\left(\frac{\partial f}{\partial x}+i\frac{\partial f}{\partial y}\right)
\end{align}

We note that the $\overline{\mathbb{R}}$-derivative is equal to zero for any holomorphic function. Recall the Cauchy-Riemann equations which state that for the complex derivative of $f(z)=g(x,y)=u(x,y)+iv(x,y)$ to exist, the following identities must hold:
\begin{align}
\frac{\partial u}{\partial x}=\frac{\partial v}{\partial y}\quad\text{and}\quad\frac{\partial v}{\partial x}=-\frac{\partial u}{\partial y}
\end{align}
If we expand the right had side of the $\mathbb{\overline{R}}$-derivative and substitute the Cauchy-Riemann equations the $\mathbb{\overline{R}}$-derivative vanishes. Thus an equivalent (and intuitive) statement about a holomorphic function is that it does not depend on the conjugate of its input. As an example, consider the loss function in Eq.~\eqref{eqn:obje}. It is real-valued and therefore non-holomorphic and it clearly depends on the conjugate of its input variable, having $\mathbb{R}$- and $\overline{\mathbb{R}}$-derivatives of $\overline{z}$ and $z$, respectively.

It is further shown by Brandwood that if $f:\mathbb{C}\to\mathbb{R}$ is a real-valued function, either $\frac{\partial f}{\partial z}=0$ or $\frac{\partial f}{\partial\overline{z}}=0$ is a necessary and sufficient condition for $f$ to have a stationary point. By extension if $f:\mathbb{C}^N\to\mathbb{R}$ is a real-valued function of a complex vector $\bm{z}=(z_1,z_2,\dots,z_N)^T$ and we define the cogradient and conjugate cogradient
\begin{align}
\nabla_{\bm{z}}&=(\partial/\partial z_1,\partial/\partial  z_2,\dots,\partial  z_N)^T\\
\nabla_{\bm{\overline{z}}}&=(\partial/\partial \overline{z}_1,\partial/\partial  \overline{z}_2,\dots,\partial\overline{z}_N)^T
\end{align}
then $\nabla_{\bm{z}}f=0$ or $\nabla_{\bm{\overline{z}}}f=0$ are necessary and sufficient to determine a stationary point. Finally, Brandwood uses Schwarz's inequality to show that the maximum rate of change of $f$ is in the direction of the conjugate cogradient $\nabla_{\bm{\overline{z}}}f$. Using these definitions, we can perform gradient descent with the conjugate cogradient operator.

\subsection{The Computational Graph}
We wish to perform gradient descent on a computational graph having a real-valued cost function and an arbitrary composition of holomorphic and non-holomorphic functions. Performing back-propagation on such a graph can be unwieldy if we choose to repeatedly switch between complex and real-valued representations of the graph. If we remain in the complex domain for all computations and use Wirtinger calculus, it is easier to build a modular framework that is useful for deep networks.

Consider a complex-valued function,
\begin{align}
\bm{F}(\bm{z},\bm{\overline{z}})&=[f_1(\bm{z},\bm{\overline{z}}), f_2(\bm{z},\bm{\overline{z}}),\dots,f_M(\bm{z},\bm{\overline{z}})]^T\quad\text{with}\\
\bm{z}&=[z_1,z_2,\dots,z_N]^T\quad\text{and}\\
\bm{\overline{z}}&=[\overline{z}_1,\overline{z}_2,\dots,\overline{z}_N]^T\quad.
\end{align}

We define the Jacobian matrices,
\begin{align}
\mathbf{J}_{\bm{F}}&\triangleq\frac{\partial \bm{F}(\bm{z,\overline{z})}}{\partial\bm{z}}\\
\mathbf{J}^c_{\bm{F}}&\triangleq\frac{\partial \bm{F}(\bm{z,\overline{z})}}{\partial\bm{\overline{z}}}
\end{align}

\begin{figure}
\begin{center}
\fbox{\includegraphics{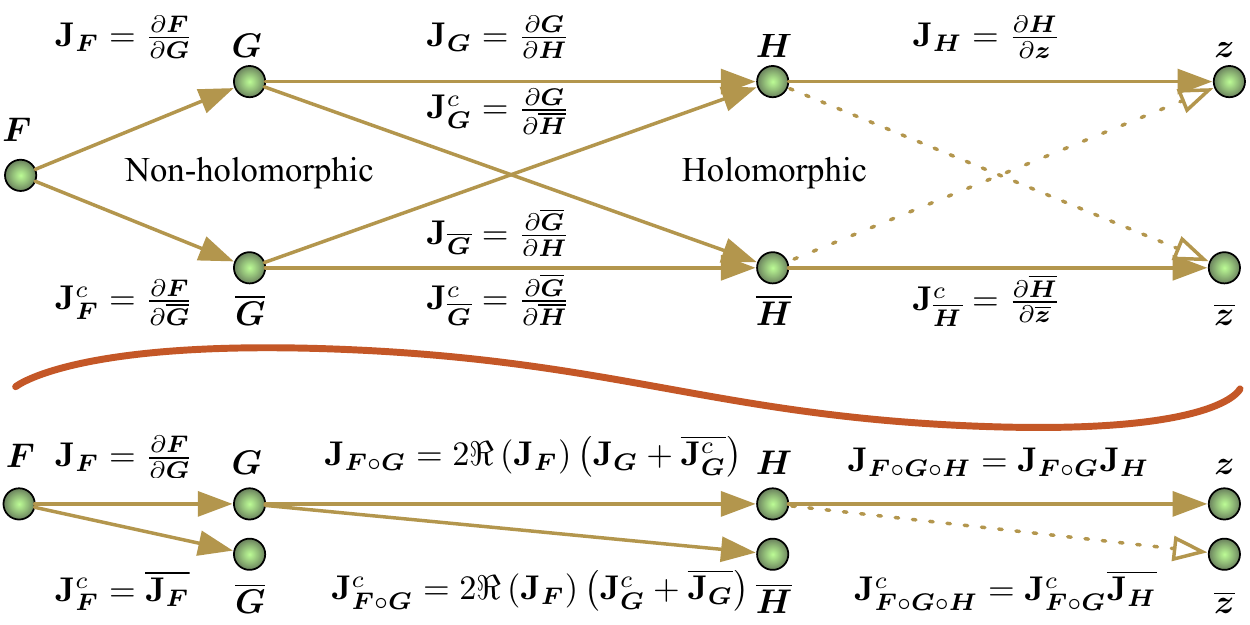}}
\end{center}
\caption{Computational dependency graph for a composition of functions $(\bm{F}\circ\bm{G}\circ\bm{H})(\bm{z,\overline{z}})$ with $\bm{F}$ real and $\bm{H}$ holomorphic. Top: naive computation requires keeping track of up to four dependencies per function. Bottom: using properties of the complex derivative we need only keep track of two dependencies per function.}\label{fig:dep}
\end{figure}

A deep neural network is constructed from a composition of several non-scalar functions. Suppose we have a composition of functions $(\bm{F}\circ\bm{G}\circ\bm{H})(\bm{z,\overline{z}})$, with $\bm{F}$ being a real-valued (non-holomorphic cost function), $\bm{G}$ being a non-holomorphic complex-valued function, and $\bm{H}$ being a holomorphic function. We would like to compute the gradient of $\bm{F}$ with respect to $\bm{\overline{z}}$. Figure~\ref{fig:dep} shows the dependency graph for back-propagating the gradient. The top part of the figure shows the Jacobian matrices for each stage of back-propagation. The naive method requires keeping track of four dependencies for every function in the graph.

We need keep track of only two partial derivatives for each function, as shown in the bottom part of Figure~\ref{fig:dep}~\citep{amin2011a,li2008a}. Keeping in mind that $\bm{F}$ is a real-valued function of complex variables and that $\bm{H}$ is holomorphic s.t. $\frac{\partial\bm{H}}{\partial\bm{\overline{z}}}=0$, we apply the chain rule to the Jacobian matrices:
\begin{align}\label{eq:jac}
\mathbf{J}^c_{\bm{F}\circ\bm{G}}&=\mathbf{J}_{\bm{F}}\mathbf{J}^c_{\bm{G}}+\mathbf{J}^c_{\bm{F}}\overline{\mathbf{J}_{\bm{G}}}\notag\\
&=2\Re\left(\mathbf{J}_{\bm{F}}\right)\left(\mathbf{J}^c_{\bm{G}}+\overline{\mathbf{J}_{\bm{G}}}\right)\\
\mathbf{J}^c_{\bm{F}\circ\bm{G}\circ\bm{H}}&=\mathbf{J}^c_{\bm{F}\circ\bm{G}}\overline{\mathbf{J}_{\bm{H}}}
\end{align}

More generally, given arbitrary functions $\bm{F}$ and $\bm{G}$ in the computational graph, we compose their Jacobians in the following way~\citep{kreutz-delgado2009a}:
\begin{align}\label{eq:jac}
\mathbf{J}_{\bm{F}\circ\bm{G}}&=\mathbf{J}_{\bm{F}}\mathbf{J}_{\bm{G}}+\mathbf{J}^c_{\bm{F}}\overline{\mathbf{J}^c_{\bm{G}}}\notag\\
\mathbf{J}^c_{\bm{F}\circ\bm{G}}&=\mathbf{J}_{\bm{F}}\mathbf{J}^c_{\bm{G}}+\mathbf{J}^c_{\bm{F}}\overline{\mathbf{J}_{\bm{G}}}\notag
\end{align}

\section{Experiments}\label{sec:expe}
The Discrete Fourier Transform of $N$ regularly-sampled points on a waveform yields complex coefficients of $N$ orthogonal complex sinusoids. However the Fourier representation may not be the best transform for a given learning task. Deep networks are regularly trained to learn data representations that are more suitable than hand-picked features. In this experiment, we generate real and complex-valued waveforms having wide-band spectral components at multiple phases and magnitudes. We train recurrent complex- and real-valued models to predict the $N$-th frame of a waveform given the first $N-1$ frames. In the following subsections we detail the data, models, and results.

\subsection{Data}
We generated four synthetic datasets having wide-band frequency spectra with random phases: Sawtooth-Like, Sawtooth-Like (Analytic), Inharmonic, and Inharmonic (Analytic). Each dataset had unique training, validation, and testing partitions. The training sets consisted of 10,000 observations split into 10 batches. The validation and test sets each had 1 batch of 1,000 observations.

Datasets were generated as described below. Each observation (waveform) has 1024 samples with a Nyquist frequency denoted $\Omega$. The waveform was split into four  non-overlapping rectangular-windowed frames of 256 samples. The first three  frames were used as input to the model and the remaining frame is reserved as ground truth for inference.

\subsubsection{Sawtooth-Like}
Each waveform has a fundamental frequency drawn uniformly from the range $[0, \Omega)$ There are harmonics $n = (2,\dots, N)$ above the fundamental frequency, with all harmonic frequencies being less than $\Omega$ and each harmonic having an amplitude of $1/n$. All sinusoidal components have a random phase drawn from a uniform distribution $[0, 1)$. Each real-valued waveform is made complex by adding a zero-valued imaginary component.

We refer to these waveforms as ``Sawtooth-Like'' because they have the same spectral components of a a band-limited sawtooth waveform. However since the phases of the spectral components are scrambled, the time-amplitude waveforms do not look like sawtooth waveforms. Each frame of an observation has a number of waveform periods in the range of $[0, 128]$. The expected number of periods per frame is 64. 

\subsubsection{Sawtooth-Like (Analytic)}
Waveforms were generated as above, but with the following modification. For each frequency component with frequency $\omega$ and phase $\phi$, a sinusoidal component is added to the imaginary axes having frequency $\omega$ and phase $\phi - \pi/2$. An analytic signal encodes instantaneous magnitude and phase. In cases where a real-valued network was trained with this dataset, the real and imaginary parts of the data were split and hence there were twice the number of inputs and outputs as other experiments.

\subsection{Inharmonic}
Inharmonic waveforms were generated with five spectral components, each having a frequency drawn from a uniform distribution in the range $[0, \Omega)$, a phase drawn from a uniform distribution in the range $[0, 1)$, and an amplitude of $1/5$. Hence the phases of the individual components are random but not drawn from the full available range of $[0, 2\pi)$. These waveforms are unlikely to exhibit periodicity.

\subsection{Inharmonic (Analytic)}
Analytic waveforms were generated as above using the same methodology as for the Sawtooth-Like (Analytic) dataset.

\subsection{Models}
We trained real- and complex-valued neural networks having one hidden recurrent layer of size 256. The input and output layers had 256 units each, with the exception of real-valued networks trained on the analytic datasets; these had 512 inputs and outputs. All models had weights and biases. Hence models had either 197,376 or 328,704 trainable parameters. Models were trained with a $\tanh$ activation function on the hidden layer and linear activation on the output layer.

\subsection{Training}
Training was performed for exactly 1000 epochs using mini-batch stochastic gradient descent with a momentum of $0.9$ and the mean squared cost function. We employed a learning rate with power scheduling decay~\citep{senior2013a}.

We and other authors have found that complex-valued networks are extremely sensitive to initial conditions and learning rates~\citep{zimmermann2011a}. In order to facilitate finding a good setting of hyperparameters, we performed hyperparameter optimization using Spearmint~\citep{snoek2012a} for the following parameters: initial weight scaling, learning rate, and learning rate decay half life. For each dataset, 100 real- and complex-valued models were trained with unique hyperparameter settings and initial weights. The final model was chosen using the best performance on the validation set.

\subsection{Results}
\begin{table}
\caption{Test Error}\label{tab:results}
\begin{center}
\begin{tabular}{l|ll}
Dataset & Complex & Real\\
\hline
\hline
Sawtooth-Like & 0.1179 & 0.1060\\
Sawtooth-Like (Analytic) & 0.3497 & 0.1937\\
Inharmonic & 0.1664 & 0.1376\\
Inharmonic (Analytic) & 0.2011 & 0.1999
\end{tabular}
\end{center}
\end{table}

\subsubsection{Overall Comparison}
Each dataset was trained, validated, and tested on a complex- and real-valued network. We had hoped that complex-valued networks would outperform real-valued nets. In most cases, the final error between complex and real nets was comparable. However in all experiments, the real-valued networks had a lower final test error. Table~\ref{tab:results} shows that both real and complex valued networks perform best on the Sawtooth-Like dataset. We were not surprised by this result. Considering that this dataset consists of only harmonically related spectral components, we presume that this dataset is easier to learn than the Inharmonic datasets.

We were surprised that both real and complex-valued networks had difficulty learning the Analytic datasets. These datasets encode instantaneous frequency and phase, and we therefore expected that they would work well with the complex valued network. It is possible that the fully complex $\tanh$ activation function is inappropriate for this dataset since instantaneous frequency does not change between inputs and outputs. In future work we will examine the performance of other activation functions on this dataset.

\subsubsection{Optimization}
\begin{figure}
\begin{center}
\includegraphics[width=0.5\linewidth]{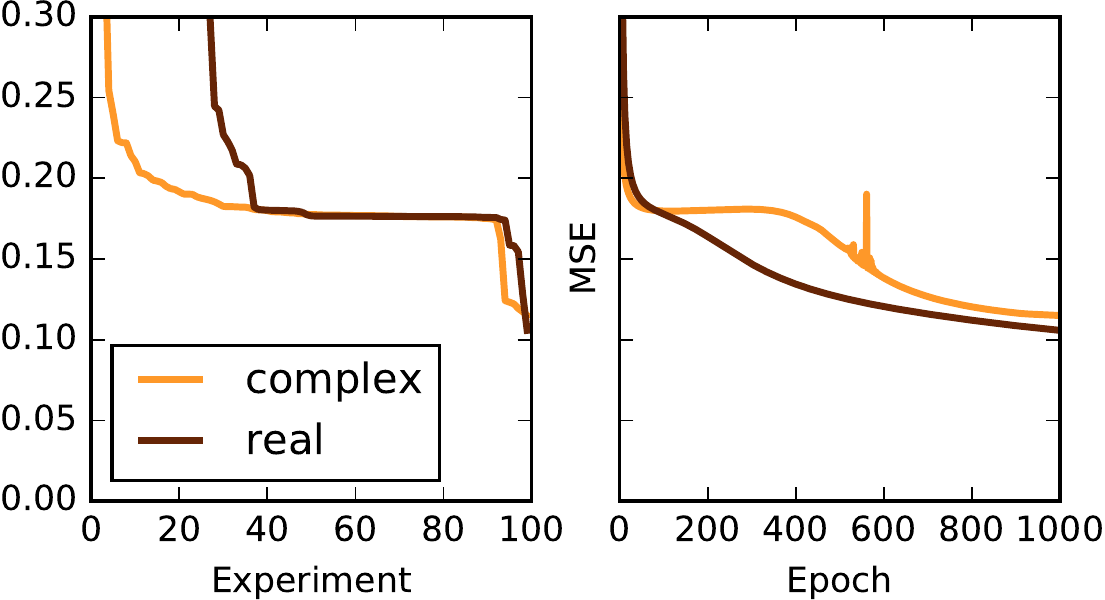}
\end{center}
\caption{Left: Sorted validation error for hyperparameter optimization. Right: Validation error for best-performing hyperparameter setting. Both figures associated with Sawtooth-Like dataset.}\label{fig:curve}
\end{figure}

The left pane of Figure~\ref{fig:curve} shows the sorted error across hyperparameter settings employed with the Sawtooth-Like dataset. We find it notable that most settings perform relatively poorly. There were only a few settings for both types of networks that achieved optimal performance. This figure underscores how sensitivities both types of networks are to hyperparameter settings.

The right pane shows the validation error across epochs for the Sawtooth-Like dataset. Note the discontinuity in the error curve for the complex-valued net. The complex valued nets are quite difficult to train and can easily approach regions of instability. We believe this is due to the singularities of the $\tanh$ function.

\subsubsection{Filters}
\begin{figure}
\begin{center}
\begin{subfigure}{0.45\textwidth}
    \includegraphics[width=\textwidth]{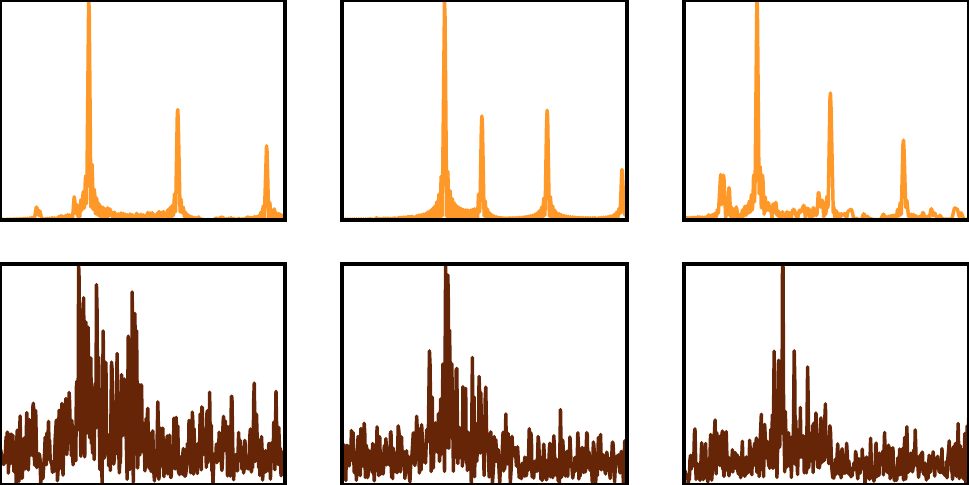}
    \caption{Sawtooth-Like (Analytic)}
\end{subfigure}
\qquad
\begin{subfigure}{0.45\textwidth}
    \includegraphics[width=\textwidth]{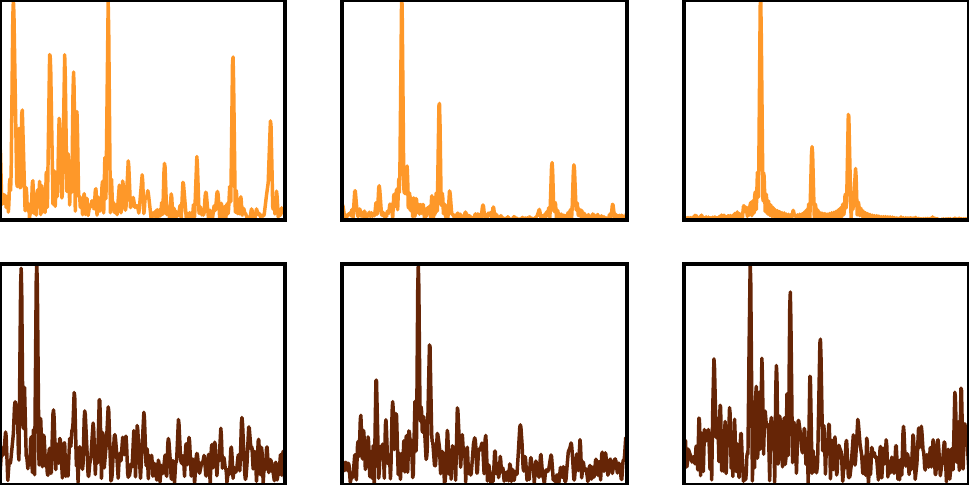}
    \caption{Inharmonic}
\end{subfigure}
\caption{Magnitude frequency response of first three filters for complex (top) and real (bottom) valued nets. The x axis shows frequency index and the y axis shows magnitude.}\label{fig:filt}
\end{center}
\end{figure}

We examined the input-to-hidden weights of the models. We found that despite the worse performance of complex-valued networks, they learned filters that are easily relatable to the datasets. Figure~\ref{fig:filt} shows the magnitude frequency responses of the first three input-to-hidden weights for the Sawtooth-Like (Analytic) (left) and Inharmonic (right) datasets. Observe that the frequency response of the complex model for the Sawtooth-Like dataset exhibits harmonically spaced peaks in the spectrum. The filters from the real-valued network are much noisier and it is difficult to discern any harmonic spacing. The filters of the complex model trained on the Inharmonic dataset also show high selectivity for a few spectral peaks, whereas, the filters learned by the real-valued model show selectivity but to a more limited degree.

\section{Conclusions}\label{sec:concl}
Despite potentially widespread applicability to machine learning tasks, the deep learning and representational learning communities have not fully embraced complex-valued nets. We argue that the mathematical conveniences of Wirtinger calculus offer a means for building a modular library for training complex-valued nets. Towards this end, we composed several synthetic datasets and compared the performance of complex- and real-valued nets. We found that complex-valued nets performed about as well as, but not better than, real-valued counterparts. We highlighted the fact that training complex-valued nets brings different challenges, including difficulties of boundedness and singularities in the activation functions. Finally we showed that despite poorer performance, complex-valued nets learn filter representations that are adapted to the domain of the data.

It is obvious that there are many challenges to successfully training complex-valued nets. We must find good methods for avoiding the singularities in holomorphic cost functions. There is no complex equivalent to the rectified linear unit. The models are extremely sensitive to initial conditions of the weights and to the learning rate. We will continue to explore these topics in future work. Our experiments were conducted on GPUs using a modified branch of the Chainer deep learning framework.\footnote{\href{http://docs.chainer.org/en/stable/index.html}{http://docs.chainer.org/en/stable/index.html}}. As we continue to investigate complex-valued networks, we intend to develop our framework further and release it to the community.

\bibliographystyle{iclr2016_conference}
\bibliography{sarroff_iclr2016}

\end{document}